\documentclass[
twocolumn,
]{ceurart}

\usepackage{algorithm}
\usepackage{algpseudocode}
\usepackage{listings}

\begin{document}

\copyrightyear{2024}
\copyrightclause{Copyright for this paper by its authors.
  Use permitted under Creative Commons License Attribution 4.0
  International (CC BY 4.0).}

\conference{KiL'24: Workshop on Knowledge-infused Learning co-located with 30th ACM KDD Conference,
August 26, 2024, Barcelona, Spain}

\title{Towards Infusing Auxiliary Knowledge for Distracted Driver Detection}


\author[1]{Ishwar B Balappanawar}[%
email=ishwar.balappanawar@students.iiit.ac.in,
url=,
]

\author[1]{Ashmit Chamoli}[%
orcid= ,
email=ashmit.chamoli@students.iiit.ac.in,
url=,
]

\author[2]{Ruwan Wickramarachchi}[%
orcid=,
email=ruwan@email.sc.edu,
url=,
]

\author[1]{Aditya Mishra}[%
orcid=,
email=aditya.mishra@students.iiit.ac.in,
url=,
]

\author[1]{Ponnurangam Kumaraguru}[%
orcid=,
email=pk.guru@iiit.ac.in,
url=,
]

\author[2]{Amit Sheth}[%
orcid=,
email=amit@sc.edu,
url=,
]
\address[1]{International Institute of Information Technology, Hyderabad}
\address[2]{AI Institute, University of South Carolina, Columbia, SC}


\begin{abstract}
Distracted driving is a leading cause of road accidents globally. Identification of distracted driving involves reliably detecting and classifying various forms of driver distraction (e.g., texting, eating, or using in-car devices) from in-vehicle camera feeds to enhance road safety. This task is challenging due to the need for robust models that can generalize to a diverse set of driver behaviors without requiring extensive annotated datasets. In this paper, we propose \textbf{KiD3},  a novel method for distracted driver detection (DDD) by \textit{infusing} auxiliary knowledge about semantic relations between entities in a scene and the structural configuration of the driver's pose.  Specifically, we construct a unified framework that integrates the scene graphs, and driver’s pose information with the visual cues in video frames to create a holistic representation of the driver's actions. Our results indicate that KiD3 achieves a 13.64\% accuracy improvement over the vision-only baseline by incorporating such auxiliary knowledge with visual information. The source code for KiD3 is available at: \url{https://github.com/ishwarbb/KiD3}.
\end{abstract}


\maketitle

\section{Introduction}

Distracted driving is a leading cause of road accidents globally, posing significant challenges to road safety. According to the National Highway Traffic Safety Administration (NHTSA)\footnote{https://www.nhtsa.gov/speeches-presentations/distracted-driving-event-put-phone-away-or-pay-campaign} approximately 3,308 people lost their lives in the United States in 2022 due to distracted driving, and nearly 290,000 people were injured. Almost 20\% of those killed in distracted driving-related crashes were pedestrians, cyclists, and others outside the vehicle. In addition to the loss of lives and injuries, the financial burden from distracted driving crashes collectively amounts to \$98 billion in 2019 alone, highlighting the urgency of developing effective detection methods.
\vspace{0.5em}

The task of identifying distracted driving involves reliably detecting and classifying various forms of driver distraction, such as texting, eating, or using other objects/devices from in-vehicle camera feeds. This task is challenging due to the need for robust models that can generalize to a diverse set of driver behaviors without requiring extensive annotated datasets. Traditionally, the DDD task has been solved using various end-to-end learning and computer vision techniques, including, but not limited to, object detection, pose estimation, and action recognition. On the other hand, recent advancements in knowledge infusion \cite{shadesofsheth2019} and Neurosymbolic AI \cite{shethneuro2023} provide new opportunities for challenging tasks in scene understanding \cite{wickramarachchi2021knowledge, wickramarachchi2022knowledge, Wickramarachchi_Henson_Sheth_2023} and context understanding \cite{oltramari2020neuro}. Hence, we posit that there is valuable auxiliary knowledge that can be either computed/ derived from the visual inputs. Specifically, we hypothesize that by infusing such knowledge with current computer vision models would improve the overall detection capabilities and robustness while not requiring the heavy computation demands of ultra-high parameter models.
\vspace{0.5em}

\begin{figure*}[t] \centering
    \newcommand{\hwidth}{0.5pt}
    \makebox[0.24\textwidth]{\small Sampled Frame}
    \makebox[0.24\textwidth]{\small Pose Estimation}
    \makebox[0.24\textwidth]{\small Object Information} 
    \makebox[0.24\textwidth]{\small Scene Graph Information} 
    \\
    \includegraphics[width=0.24\textwidth]{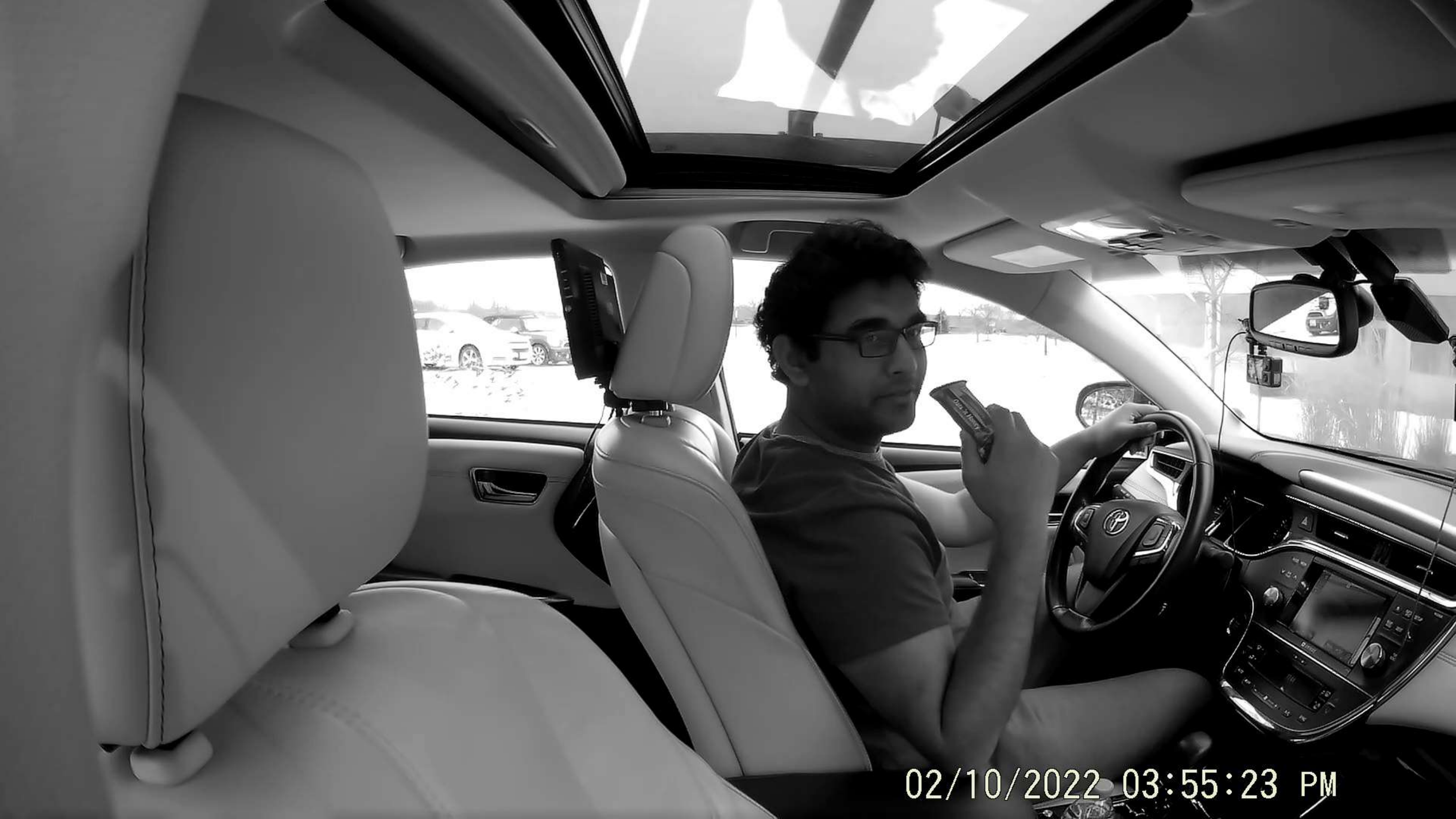}
    \includegraphics[width=0.24\textwidth]{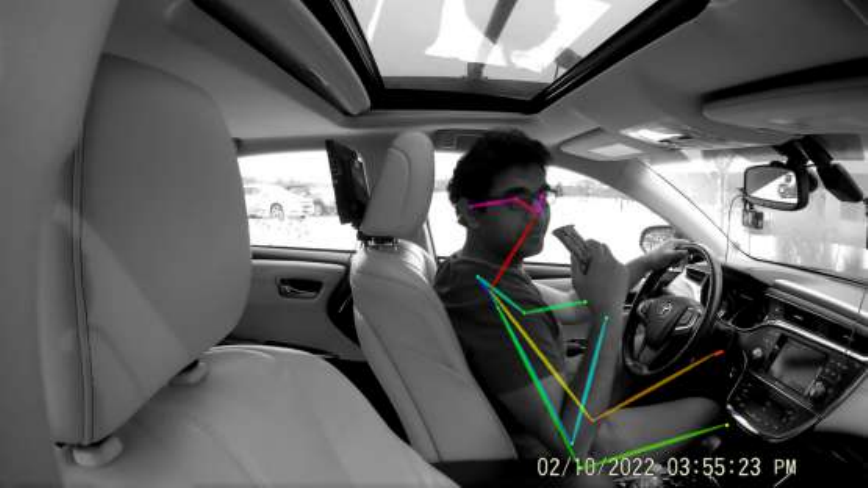}
    \includegraphics[width=0.24\textwidth]{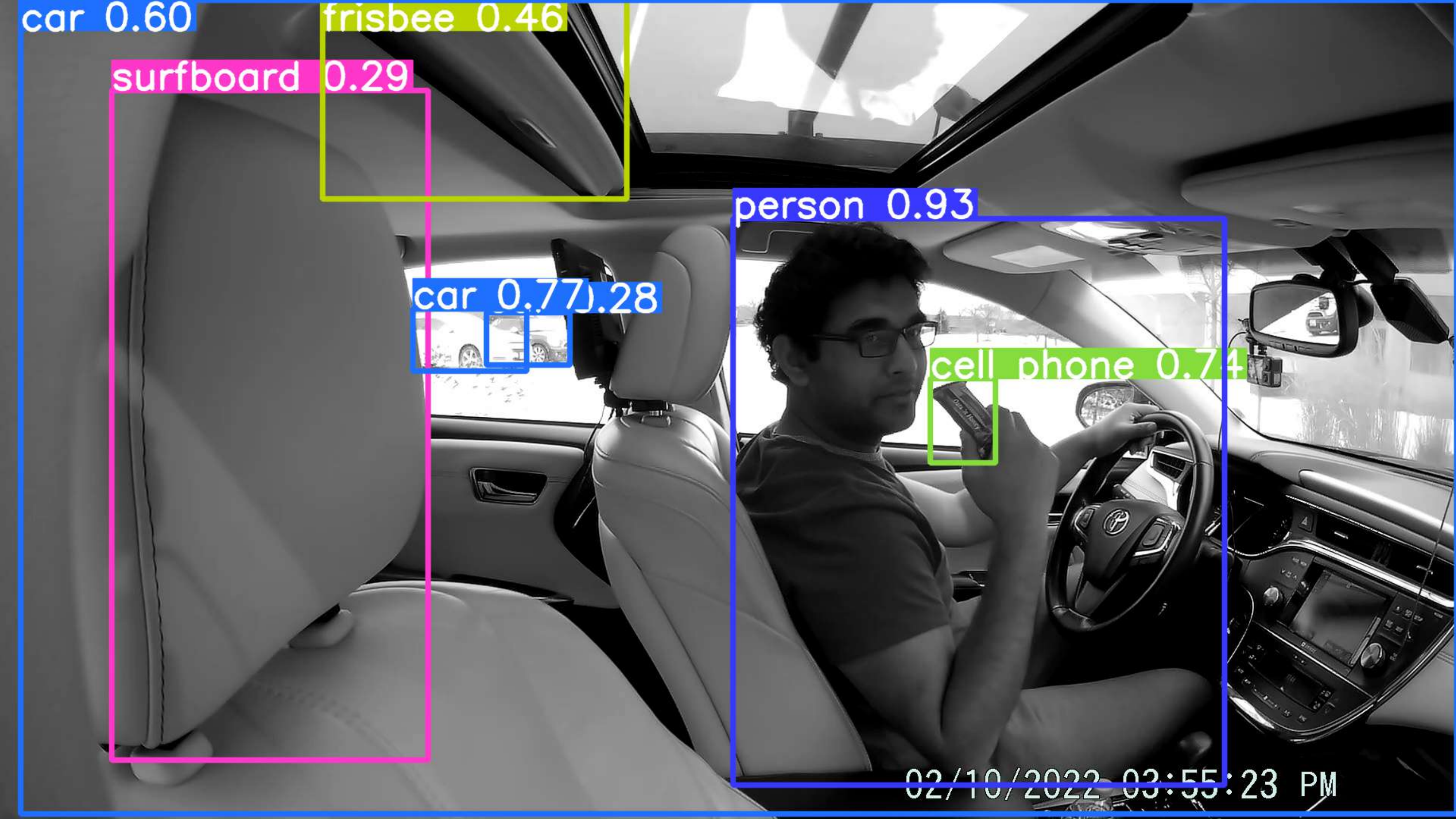}
    \includegraphics[width=0.24\textwidth]{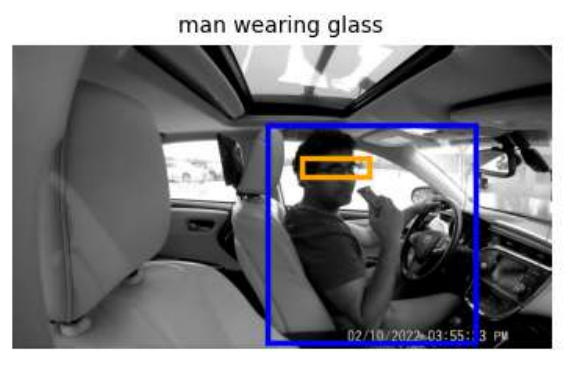}
    \caption{This figure illustrates the process of extracting detailed information from a scene to analyze driver behavior. The extreme left panel shows an image of a driver which is sampled from the video. The middle left panel presents the corresponding estimated pose, highlighting how structured representations can be derived from raw image data. The middle right panel presents the object information obtained via object detection.The extreme right panel provides an sample relation from the scene graph, capturing the relationships between different objects and actions.  }
    \label{fig:process-workflow}
\end{figure*}

To this end, we propose \textbf{KiD3}, a novel, simplistic method for distracted driver detection that infuses auxiliary knowledge about inherent semantic relations between entities in a scene and the structural configuration of the driver's pose. Specifically, we construct a unified framework that integrates scene graphs and the driver’s pose information with visual information to enhance the model's understanding of distraction behaviors (see Figure \ref{fig:process-workflow}). 
\vspace{0.5em}

Conducting experiments on a real-world, open dataset, our results indicate that incorporating such auxiliary knowledge with visual information significantly improves detection accuracy. KiD3 achieves a 13.64\% accuracy improvement over the vision-only baseline, demonstrating the effectiveness of integrating semantic and pose information in DDD tasks. This improvement highlights the potential of our method to contribute to safer driving environments by providing a more reliable, efficient and scalable solution that does not demand the use of expensive high-parameter models.

\noindent Contributions of this paper are as follows:

\begin{enumerate}
    \item A novel, simple method for distracted driver detection that incorporates the auxiliary knowledge computed/estimated with vision inputs without the need for high-parameter, computational heavy models.
    \item A demonstration of the effectiveness of infusing different types of auxiliary knowledge over vision-only baselines using real-world distracted driving data.
    
\end{enumerate}

\section{Related Work}
{Distracted Driver Detection} is generally formulated as one of 2 tasks: \textbf{Action Recognition/Classification} and \textbf{Temporal Action Localization} (TAL). Action recognition is a computer vision task that involves classifying a given image or a video into a set of pre-defined set of actions or classes. TAL, on the other hand detects activities being performed in a video streams and outputs start and end timestamps. In this paper, we focus on solving the action recognition task by classifying frames into various distracted driver activities. Here, we explore related work considering two directions: (1) methods for distracted driver identification and (2) methods for generating/encoding semantic graphs from visual scenes.\\

\noindent \textbf{Existing Methods for DDD:} Vats et al.\cite{key-point} proposes Key Point-Based Driver Activity Recognition that extracts static and movement-based features from driver pose and facial features and trains a frame classification model for action recognition. Then, a merge procedure is used to identify robust activity segments while ignoring outlier frame activity predictions.
\vspace{0.5em}

In their work, Tran et al. \cite{tran} utilize multi-view synchronization across videos by training an ensemble 3D action recognition model on each view and taking the average probability over all the views as the final output. The outputs are then post-processed for predicting the action label and temporal localization of the predicted action. This work utilizes the X3D family of networks \cite{DBLP:journals/corr/abs-2004-04730} for video classification instead of relying on manual feature engineering.
Wei Zhou et al. \cite{zhou} improve upon this work by fine-tuning large pre-trained models instead of training from scratch and by empirically selecting specific camera views for specific distracted action classes.
\vspace{0.5em}

Previous works mainly focus on the use of sophisticated post-processing algorithms, use of larger encoder-decoder architectures and multi-view synchronization to improve action recognition and TAL performance. In contrast, our work aims to improve classification performance by \textit{incorporating auxiliary knowledge (e.g.,  semantic entities/relationships of a frame, pose information) that can be derived and infused as graphs} into the encoder side of our architecture. Next, we will explore the state-of-the-art methods for scene graph generation.\\

\noindent\textbf{Scene Graph Generation} (SGG) refers to the task of automatically mapping an image or a video into a semantic structural scene graph, which requires the correct labeling of detected objects and their relationships \cite{zhu2022scene}. 
Yuren Cong et al. \cite{cong2023reltr} pose SGG as a set prediction problem. They propose an end-to-end SGG model, RelTR, with an encoder-decoder architecture. In contrast to most existing scene graph generation methods, such as Neural Motif, VCTree, and Graph R-CNN, \cite{Zellers_2018_CVPR, DBLP:journals/corr/abs-1812-01880, Yang_2018_ECCV} which RelTR used as benchmarks, RelTR is a one-stage method that predicts sparse scene graphs directly only using visual appearance without combining entities and labeling all possible predicates. Due to its simplicity, efficiency and SOTA performance, we selected RelTR to generate SGGs for our experiments.
\vspace{0.5em}

Additionally, inspired by the work of Pen Ping et al. \cite{PING2023121} 
we incorporate atomic action information extracted from the objects detected in the scene and the estimated pose of the driver.

\section{Methodology}
In this section, we formally define the DDD problem, the datasets used, preprocessing steps, and delve deep into the technical details of each sub-component in the proposed approach (see Figure \ref{fig:workflow}). 

\subsection{Problem Statement}
Given a video frame $\mathbf{x} \in \mathbb{R}^{m \times n \times 3}$ sampled from a video where $m$ denotes the height of the frame, $n$ denotes the width of the frame, and 3 corresponds to the color channels (RGB), the learning objective is to classify it into one of 18 predefined activities  $\mathcal{C} = \{C_1, C_2, \ldots, C_{18}\}$.
\vspace{1em}

We define a classifier model $f: \mathbb{R}^{m \times n \times 3} \to [0, 1]^{18}$ that maps a video frame to a probability distribution over the 18 activities. Specifically, $f(\mathbf{x}) = \mathbf{p}$, where $\mathbf{p} = [p_1, p_2, \ldots, p_{18}]$ and $p_i$ represents the probability that the frame $\mathbf{x}$ belongs to class $C_i$, such that $\sum_{i=1}^{18} p_i = 1 \quad \text{and} \quad 0 \leq p_i \leq 1 \quad \forall i \in \{1, \ldots, 18\}.$ The predicted class $\hat{C}$ for the frame $\mathbf{x}$ can therefore be determined by: $\hat{C} = \arg\max_{C_i \in \mathcal{C}} p_i.$

\subsection{Datasets for DDD}

\begin{figure}[t]
\centering
\includegraphics[width=0.45\textwidth]{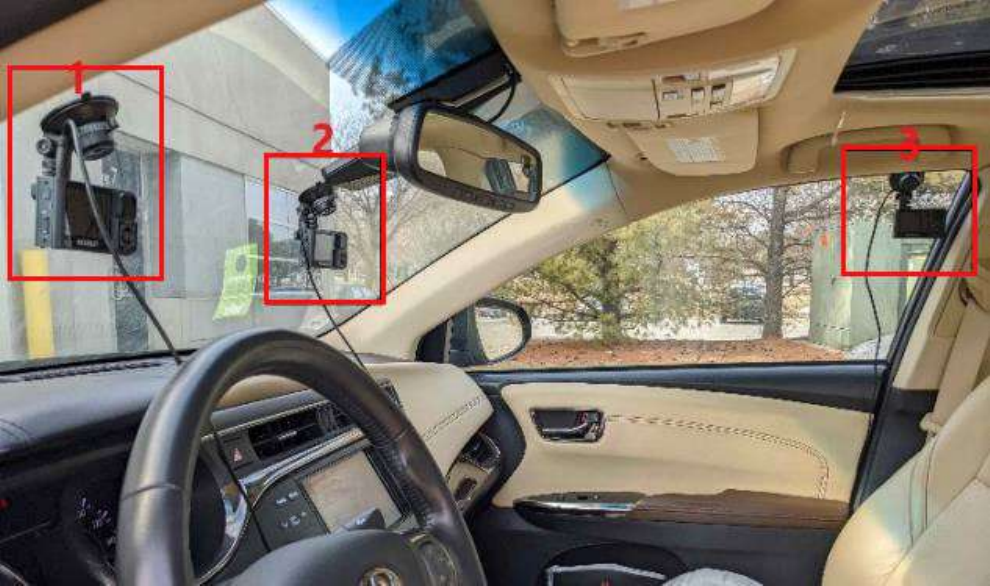}
\caption{Camera mounting setup for the three views in the {\it SynDD1} dataset: 1. Dashboard, 2. Behind rear view mirror, and 3. Top right side window.}
\label{fig:track3:camera}
\end{figure}

The real-world datasets for distracted driver identification typically include annotated video sequences from cameras mounted inside the vehicle. While several open datasets are available, such as \texttt{StateFarmDataset}\footnote{https://www.kaggle.com/competitions/state-farm-distracted-driver-detection}, we have selected SynDDv1 \cite{RAHMAN2023108793} to be used for experiments due to the higher number of distracted behavior classes and the diversity, including variations in lighting conditions, driver appearances, and the use of objects and people in the background. SynDDv1 consists of 30 video clips in the training set and 30 videos in the test set. The dataset consists of images collected using three in-vehicle cameras positioned at locations: on the dashboard, near the rear-view mirror, and on the top right-side window corner, as shown in Table 1 and Figure 1. The video sequences are sampled at 30 frames per second at a resolution of 1920×1080 and are manually synchronized for the three camera views. Each video is approximately 10 minutes long and contains all 18 distracted activities shown in Table 2. The driver executed these activities with or without an appearance block, such as a hat or sunglasses, in random order for a random duration. There are six videos for each driver: three videos with an appearance block and three videos without any appearance block.


\begin{table}[]
\caption{The list of distracted driving activities in the {\it SynDD1} dataset.}
\centering
\begin{tabular}{|c|c|}
\hline
Sr. no. & Distracted driver behavior      \\ \hline
1       & Normal forward driving            \\ \hline
2       & Drinking                          \\ \hline
3       & Phone call (right)                \\ \hline
4       & Phone call (left)                 \\ \hline
5       & Eating                            \\ \hline
6       & Texting (right)                      \\ \hline
7       & Texting (left)                       \\ \hline
8       & Hair / makeup                     \\ \hline
9       & Reaching behind                   \\ \hline
10      & Adjusting control panel           \\ \hline
11      & Picking up from floor (driver)    \\ \hline
12      & Picking up from floor (passenger) \\ \hline
13      & Talking to passenger at the right \\ \hline
14      & Talking to passenger at backseat  \\ \hline
15      & Yawning                           \\ \hline
16      & Hand on head                      \\ \hline
17      & Singing with music                \\ \hline
18      & Shaking or dancing with music     \\ \hline
\end{tabular}
\label{tab:driving:activities}
\end{table}

\begin{figure*}
  \includegraphics[width=\textwidth]{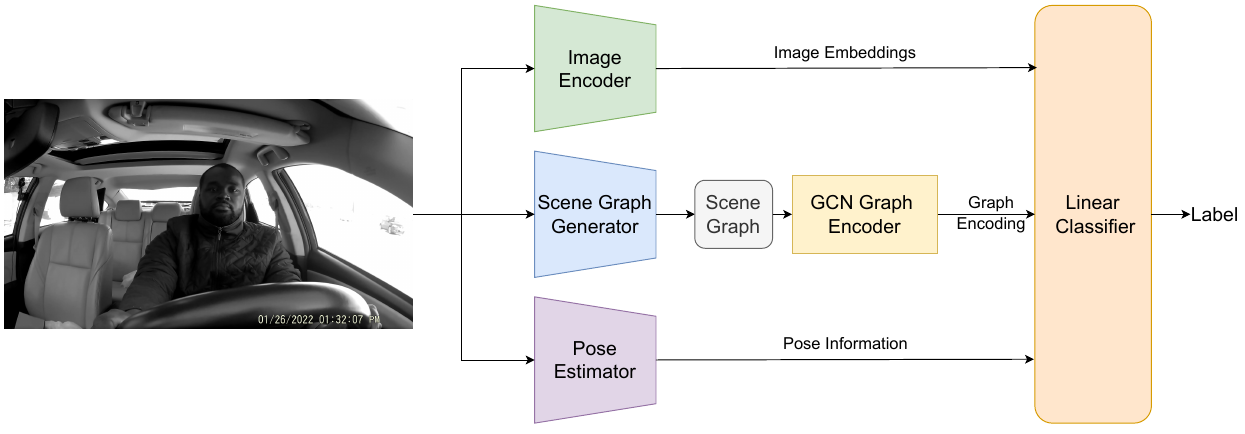}
  \caption{Workflow of our proposed method. The figure illustrates the integration of an Image Encoder, Scene Graph Generator, GCN Graph Encoder, and Pose Estimators within our pipeline.}
  \label{fig:workflow}
\end{figure*}

\subsection{Data Preprocessing}

From the dataset, we selected the Dashboard variant, resulting in 10 videos for training and 10 videos for testing. Sets of (frame, label) were created by sampling frames from the videos at regular intervals and obtaining the corresponding labels from the annotations. The publicly available dataset contains various inconsistencies in the annotation format provided as CSV files. These inconsistencies, such as different naming conventions, variations in capitalization, and extra spaces in names, have been resolved to ensure consistency across all data splits.

\vspace{0.5em}
Next, we will outline the technical details for each sub-component in our  approach, shown in Figure \ref{fig:workflow}.

\subsection{Image Encoding}

\subsubsection{Background}
To classify video frames into one of the predefined activities, the first step is to obtain robust image embeddings that would effectively capture the visual features in raw pixel data into a more manageable and informative representation. Possible methods for this transformation include using pre-trained Convolutional Neural Networks (CNNs) like VGGNet \cite{simonyan2014very}, ResNet \cite{He_2016_CVPR}, or Inception \cite{Szegedy_2015_CVPR}. Out of these methods, we selected VGG16, a variant of VGGNet, due to its simplicity and effectiveness in extracting deep features from images. VGG16 has been extensively used and validated in various image classification tasks, making it a reliable choice for our purpose.

\subsubsection{Technical Details}
VGGNet, particularly VGG16, is a deep convolutional network known for its simple yet effective architecture, consisting of 16 weight layers. The network is structured with multiple convolutional layers followed by fully connected layers. Each convolutional layer uses small receptive fields (3x3) and applies multiple filters to extract features at different levels of abstraction. The fully connected layers then process these features for classification. VGG16's design focuses on depth and simplicity, making it an ideal candidate for transfer learning.

\subsubsection{Pre-processing and Adaptation}
To adapt VGG16 for our task, we fine-tuned the model to obtain image embeddings. Specifically, we discarded the last 2 classifier layers of the pre-trained VGG16 model and retained the base model along with the first 4 classifier layers. This configuration results in a 4096-dimensional image embedding vector. The rationale for discarding the last 2 layers is that the final layer reduces the dimensionality to only 18, which is insufficient for our needs. Additionally, the earlier layers capture more general features, which are beneficial for transfer learning. These embeddings are then used for further processing and classification tasks.

\subsection{Scene Graph Generation and Encoding}

\subsubsection{Background}
Scene graphs structurally represent the relationships between various objects in a given image. Each node in the graph represents an object, while edges denote the relationships between these objects; for example consider the triple:  ``<< man \textit{holding} phone >>''. Scene graphs capture the high-level contextual and semantic information of the scene, going beyond pixel-level data. They are also essential for scene understanding and reasoning and allow us to explicitly inject knowledge into the pipeline. For example, considering DDD task, a scene graph containing the triple ``<< person \textit{drinking\_from} bottle >>'' might indicate distracted driving activity.
Modeling such important relations can otherwise be achieved \textit{implicitly} using methods such as convolutional-network-based image encoders, with some uncertainty.

\subsubsection{Technical Details}
To generate the scene graph for a given frame, we use the RelTr architecture \cite{cong2023reltr}. Then, we use a Graph Convolutional Network (GCN) \cite{kipf2017semisupervised} layer followed by a $Tanh$ activation to obtain representations for each node in the graph. We take the mean of all the node embeddings to obtain a graph-level representation and treat this vector as the graph encoding.

\subsubsection{Pre-processing and Adaptation}
A scene graph output from RelTr \cite{cong2023reltr} is in the form of triplets of the form $(node, \textit{relation}, node)$. Essentially, we get a list of relations $R_i = (n_1, \textit{r}, n_2)$ where $n_1$ and $n_2$ are nodes and $\textit{r}$ is the relation between them. This format is converted to a list of \textit{edges}, where edges are represented as pairs of nodes. This is provided to the GCN encoder to obtain a graph-level representation.

\subsection{Pose Estimation}
\subsubsection{Background}
Pose estimation is a critical component in understanding the spatial configuration of a subject's body, which in this case is the driver. By capturing the positions of key body parts, pose estimation provides valuable information about the driver’s posture and movements. This information is essential for accurately classifying the driver's activities. Various methods can be employed for pose estimation, including 2D and 3D approaches. We opted to use a state-of-the-art 2D pose estimation technique to effectively capture the required spatial data. 

\subsubsection{Technical Details}
We utilized OpenPose \cite{Cao_2017_CVPR}, a state-of-the-art 2D pose estimation model, to extract pose information. OpenPose can detect and output a set of key points corresponding to various body parts, such as the head, shoulders, elbows, and hands. These key points are represented as coordinates in a 2D space. The process involves detecting the spatial locations of these joints and constructing a pose structure that reflects the driver’s body configuration. Mathematically, each key point can be represented as: $\mathbf{k}_i = (x_i, y_i)$ where $\mathbf{k}_i$ denotes the $i$-th key point with $x_i$ and $y_i$ being its coordinates in the image frame.

\subsubsection{Pre-processing and Adaptation}
To adapt the pose estimation data for our task, we pre-processed the key point coordinates obtained from OpenPose. The key points were normalized and structured to consistently represent the driver’s pose. 

Additionally, we derived features such as the distance between the hands and eyes/face, the angle formed by the eyes with the neck, and the distance between the hands and objects like a phone or bottle (if detected using YOLO \cite{Redmon_2016_CVPR}). These features were crucial for enhancing the model's ability to accurately interpret and classify the driver's activities.

            
    

\subsection{Unified Pipeline}

We construct a simple machine-learning pipeline to combine the latent encodings of the above modules. Each module takes an image as input and processes it into a meaningful vector representation. We then concatenate these representations using a feed-forward MLP to classify the input image. Algorithm 1 succinctly outlines the main steps of this pipeline.

\begin{figure}[!ht]
\centering
\includegraphics[width=\columnwidth]{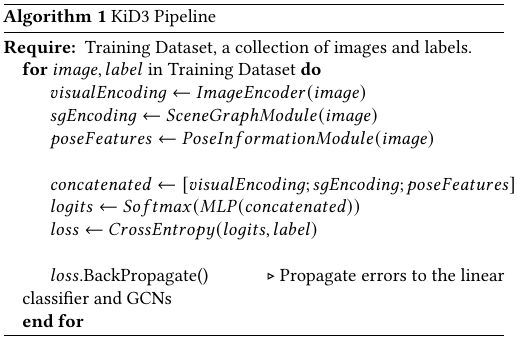}
\label{fig:kid3_pipeline}
\end{figure}

        


\subsubsection{Training}
We first fine-tune the pre-trained image encoder on the distracted driver classification task to obtain task-suitable embeddings. During training, we freeze the Image Encoding and Pose Information modules and only train the linear classifier and the GCN graph encoder in the Scene Graph Encoding module. We use $Softmax$ activation in the final layer of the feed-forward MLP and use the Cross-Entropy loss function.

\section{Experiments}

We outline the following experimental setup to evaluate the proposed approach's overall performance and the contribution of each sub-component.
    
\subsection{Method 1 - Vision Only}

In the first experiment, we utilized existing computer vision (CV) models to establish a baseline performance for the frame classification task. We fine-tuned the VGG-16 model to assess the performance of traditional CV models. To achieve this, we froze the weights of the entire model and unfroze only the classification layers (model.classifier[1...6]). The sixth classification layer \texttt{nn.Linear(4096, 1000)} was replaced with \texttt{nn.Linear(4096, 18)} to match the number of activity classes. The modified model was then fine-tuned on our classification task, allowing the classification layers to adapt to the specific features of our dataset.

\subsection{Method 2 - Vision + Scene Graphs}
In the second experiment, we use the VGG-16 similar to how it was used in Method 1; however, out of the last six classifier layers, we discarded the last two layers and used the base model with the first four classifier layers to obtain a 4096-dimensional image embedding vector. The rationale is that the final layer could not be utilized because it reduces the image embedding to only 18 dimensions, which is insufficient for capturing the rich features needed for our task. Moreover, earlier layers in the network capture more general features beneficial for transfer learning. Then, we integrate image embeddings with scene graphs encoded using a Graph Convolutional Network (GCN) \cite{kipf2017semisupervised}. The embeddings derived from the GCN are concatenated with the image embeddings obtained from the VGG-16 model. Linear layers are used as a head to combine these information streams, forming a unified representation. This combined model was trained on the same classification objective, leveraging both the visual and relational features present in the data.

\subsection{Method 3 - Vision + Scene Graphs + Pose Information}
In the final experiment, we further enrich the scene representation by incorporating pose information, enhancing its ability to understand the driver’s activities. The pose details included the location of objects via bounding boxes and the outline of the human skeleton with coordinates of key points such as the eyes, nose, and fists. We engineered additional features based on external knowledge, including the distance between the hand and face and the distance between the hand and a phone or bottle (if detected using YOLO \cite{Redmon_2016_CVPR}). These engineered features were added to the concatenation of image embeddings and scene graph embeddings. The model is then re-trained on the classification task with these additional features, providing a holistic understanding of the driver’s activities.

\begin{table*}[!ht]
  \caption{Performance of the three methods on the test set}
  \label{tab:performance}
  \begin{tabular}{cccccc}
    \toprule
    \textbf{Method} & \textbf{Accuracy} & \textbf{F1 Score} \\
    \midrule
    Vision Only & 79.64 $\pm$ 2.17\% & 0.81\\
    Vision + Scene Graphs & 89.1 $\pm$ 1.61\% $~~(\uparrow11.88\%)$ & 0.89 $~~(\uparrow9.88\%)$ \\
    Vision + Scene Graphs + Pose Information & 90.5 $\pm$ 1.32\%  $~~(\uparrow13.64\%)$& 0.91 $~~(\uparrow12.35\%)$ \\
    \bottomrule
  \end{tabular}
\end{table*}

\section{Results}
Table \ref{tab:performance} summarizes the results of our experiments on the test set and the ablation studies across different method variations. We evaluate the performance using two metrics: accuracy and the F1 score. The vision-only model achieves 79.64 overall accuracy and 0.81 F1 score, respectively. With the inclusion of scene graphs, the accuracy and the F1 score increased by 11.88\% and 9.88\%, respectively. Finally, the complete model incorporating both scene graphs and pose information achieves the peak performance of 90.5\% accuracy and 0.91 F1 score, respectively.  


\begin{figure}[!ht]
\centering
\includegraphics[width=0.5\textwidth]{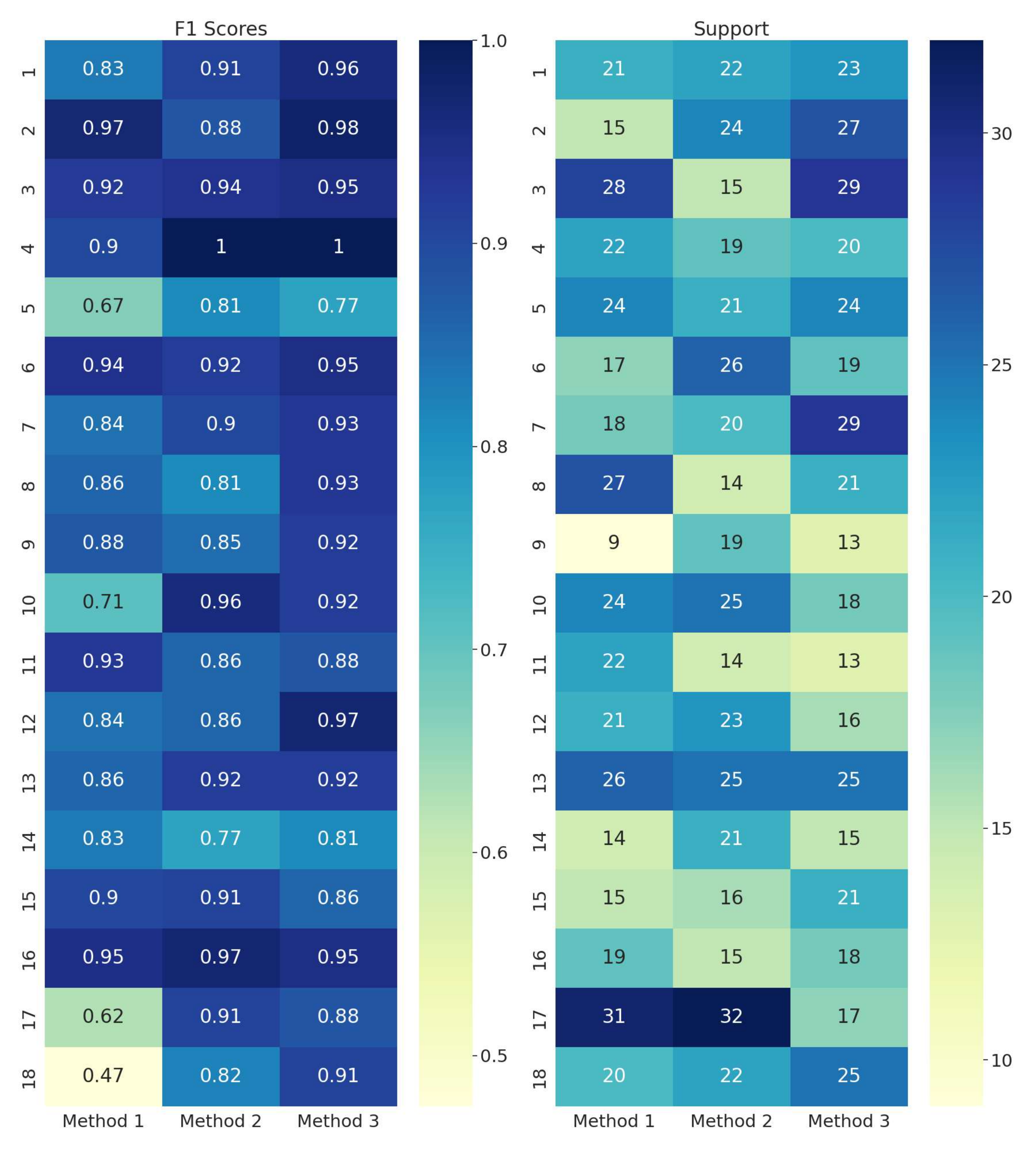}
\caption{F1 scores and support for individual activity (i.e., \textit{Class 1 - 18}) prediction across three methods, with Method 2  (i.e., Vision + SGG) and Method 3 (i.e., Vision + SGG + Pose Info) showing improvements over Method 1 (i.e., Vision only).}
\label{fig:track3:camera1}
\end{figure}

We have observed (see Figure 4) that our methods are particularly effective in identifying classes such as Eating (class 5), Adjusting Control Panel (class 10), and Singing with Music (class 17). We interpret this as evidence that our approach successfully incorporates auxiliary knowledge, enhancing our model's performance for these classes.

\section{Discussion}
Our results clearly support the initial hypothesis that the inclusion of valuable auxiliary knowledge with visual features would enhance the performance of the DDD task. The ablation study further establishes each auxiliary knowledge type's role in the overall performance. Scene graphs provided the most significant auxiliary knowledge, highlighting the importance of explicitly encoding semantic information and infusing it with visual features. By incorporating pose information of driver actions, we were able to further enrich overall accuracy and robustness. However, several limitations to our approach warrant further investigation.

\subsection{Limitations} One limitation is the reliance on annotated data for training. While we used a combination of supervised and unsupervised learning techniques to mitigate this issue, the availability of annotated data remains a key constraint. Additionally, our method may struggle with complex and highly variable driving scenarios where the relationships between objects and actions are less clear. Finally, we have not considered using foundation models like Vision Language Models (VLMs) for our experiments. Our main focus in this work is to evaluate the impact of auxiliary knowledge on the DDD task without the need for complex, high-parameter models. 

\section{Conclusions and Future Work}

In this paper, we proposed a novel, simple approach to distracted driver detection by infusing two types of auxiliary knowledge with visual information. Our method leverages scene graphs and estimated pose information with visual embeddings to comprehensively represent driver actions. Our experimental results showcase the effectiveness of infusing each type of auxiliary knowledge with visual features to achieve 90.5\% peak performance on the DDD task.

Future work will address the limitations mentioned above, such as the reliance on annotated data and the handling of complex driving scenarios. Additionally, we plan to explore the integration of other types of knowledge representations, such as temporal graphs, to further enhance the performance of distracted driver detection systems Further, we plan to investigate the role of VLMs in this task.


\end{document}